\documentclass{article}

\usepackage[preprint,nonatbib]{neurips_2019}
\usepackage[utf8]{inputenc} % allow utf-8 input
\usepackage[T1]{fontenc}    % use 8-bit T1 fonts
\usepackage{url}            % simple URL typesetting
\usepackage{amsmath,amsfonts,amsthm}       % blackboard math symbols
\usepackage{nicefrac}       % compact symbols for 1/2, etc.
\usepackage{microtype}      % microtypography
\usepackage{graphicx}
\usepackage{array}
\usepackage{multirow}
\usepackage{adjustbox}
\usepackage{mathptmx}
\usepackage{hyperref}
\usepackage{verbatim}
\usepackage{subfigure}
\usepackage{tabu,booktabs}
\usepackage{caption}
\usepackage{mathtools}
\usepackage{setspace}
\usepackage{mwe}
\usepackage[backend=bibtex,style=alphabetic,maxbibnames=99]{biblatex}
\DeclareMathOperator*{\argmin}{arg\,min}
\newtheorem{theorem}{Theorem}
\newtheorem{lemma}[theorem]{Lemma}

\title{Weight Friction: A Simple Method to Overcome Catastrophic Forgetting and Enable Continual Learning}

\author{%
  Gabrielle K. Liu\\
  \texttt{gkml@mit.edu}\\
  }

\begin{document}

\maketitle

\begin{abstract}
In recent years, deep neural networks have found success in replicating human-level cognitive skills, yet they suffer from several major obstacles. One significant limitation is the inability to learn new tasks without forgetting previously learned tasks, a shortcoming known as catastrophic forgetting. In this research, we propose a simple method to overcome catastrophic forgetting and enable continual learning in neural networks. We draw inspiration from principles in neurology and physics to develop the concept of weight friction. Weight friction operates by a modification to the update rule in the gradient descent optimization method. It converges at a rate comparable to that of the stochastic gradient descent algorithm and can operate over multiple task domains. It performs comparably to current methods while offering improvements in computation and memory efficiency.
\end{abstract}

%%%%%%%%%%%%%%%%%%%%%%%%%%%%%
%% INTRODUCTION
%%%%%%%%%%%%%%%%%%%%%%%%%%%%%
\section{Introduction}\label{sec1}

In recent years, deep neural networks have found success in various applications of artificial intelligence~\cite{lecun2015deep,deng2014deep,schmidhuber2015deep}. They have achieved vast improvements in areas ranging from speech recognition to cancer detection. These and other benefits of deep learning can be expected to provide significant improvements to the quality of human life. While neural networks hold great promise for achieving human-level intelligence, there remain several fundamental obstacles to replicating human cognitive skills and achieving strong AI. 

One of the most significant limitations is that neural networks lack the ability to continually learn: they struggle to retain previously acquired knowledge and experience after learning to perform new tasks. The ability to perform continual learning is essential for computational learning systems to achieve human-level intelligence. To realize continual learning in neural networks, we must overcome a key shortcoming known as catastrophic forgetting.

Catastrophic forgetting refers to when a neural network cannot learn tasks sequentially without ``forgetting’’ how to perform previously learned tasks~\cite{goodfellow2013empirical}. It arises as a consequence of a neural network's inability to strike a balance between plasticity (the ability to adapt to new tasks and learn new information) and stability (the ability to preserve previously learned important information)~\cite{robins1995catastrophic}. This phenomenon is known as the \textit{stability-plasticity dilemma}. Ideally, neural networks should be able to generalize and apply previous knowledge to new tasks by learning representations that are applicable across a wide variety of domains. In reality, the excessive plasticity of neural networks leads to catastrophic forgetting, which prevents continual learning. 

In this paper, we propose weight friction as an simple, effective, and efficient method to address this challenge. We develop an update rule for gradient descent in neural networks that allows weight values to become more resistant to change (less plasticity) as the network learns new tasks. In this way, it becomes possible for neural networks to overcome catastrophic forgetting and thereby learn in a continual fashion. 

The remainder of this paper is organized as follows. In Section \ref{sec2}, we briefly discuss related work. In Section \ref{sec3.1}, we introduce the motivation and intuition behind the concept of weight friction. In Section \ref{sec3.2}, we analyze the rate of convergence for training neural networks with weight friction. In Section \ref{sec3.3}, we evaluate the effect of weight friction on several continual learning settings. In Section \ref{sec4}, we discuss the relevance and importance of our results. Finally, in Section \ref{sec5}, we review our conclusions and discuss avenues for future work.

%%%%%%%%%%%%%%%%%%%%%%%%%%%%%
%% RELATED WORK
%%%%%%%%%%%%%%%%%%%%%%%%%%%%%
\section{Related work}\label{sec2}

Catastrophic forgetting has been a well-known problem in neural networks since the 1980s~\cite{kirkpatrick2017overcoming}. Many approaches to overcome catastrophic forgetting and facilitate transfer learning have been proposed. In general, these methods exhibit issues such as computational complexity, poor scalability, increased training time, and difficulty of implementation. One approach, called Progressive Neural Networks (PNNs), instantiates a new neural network for each task in consideration and uses lateral connections between the hidden layers of the networks to transfer knowledge~\cite{rusu2016progressive}. The creation of a separate neural network for each task significantly increases computation time and memory use, thus making PNNs not easily scalable to a large number of tasks. Another method, Gradient Episodic Memory (GEM), uses an episodic memory for each task in consideration and avoids catastrophic forgetting by minimizing both the loss for each task and the losses on the episodic memories for previous tasks~\cite{lopez2017gradient}. While GEM yields appreciable gains in performance, it also results in a large computational burden during training. To address this problem, a method has been proposed that minimizes the average episodic memory loss rather than the episodic memory loss for each previous task \cite{chaudhry2018efficient}. This modified version of GEM is known as Averaged GEM (A-GEM) and is significantly more computationally efficient but still requires an episodic memory to be maintained. Elastic Weight Consolidation (EWC) is a method that seeks to overcome catastrophic forgetting by regularizing the loss with importance parameters based on a Laplace approximation to the Fisher information matrix calculated at the end of each task~\cite{kirkpatrick2017overcoming}. Another method applies the Benna-Fusi model to perform Weight consolidation over a range of time scales \cite{kaplanis2018continual}, but it requires storing the time and magnitude of every parameter update. We offer weight friction as a method to overcome catastrophic forgetting that avoids many of the aforementioned issues while offering substantial improvements in computational efficiency.

%%%%%%%%%%%%%%%%%%%%%%%%%%%%%
%%%%%%%%%%%%%%%%%%%%%%%%%%%%%
%%%%%%%%%%%%%%%%%%%%%%%%%%%%%
%%%%%%%%%%%%%%%%%%%%%%%%%%%%%
\section{Weight friction}
%%%%%%%%%%%%%%% MOTIVATION %%%%%%%%%%%%%%%
\subsection{Motivation}\label{sec3.1}
We propose a mechanism for avoiding catastrophic forgetting in neural networks that is analogous to a mechanism in the human brain and the concept of friction in physics. There is evidence that the human brain can prevent catastrophic forgetting by protecting knowledge in neocortical circuits \cite{segal2010dendritic}. When learning occurs, a portion of the excitatory synapses in the brain is strengthened, and there is a resulting enlargement of dendritic spines (small protrusions covering the surface of a neuron's dendrites that receive input from excitatory synapses) \cite{segal2010dendritic, nimchinsky2002structure}. Compared to small spines, which are transient and easily erased, enlarged spines are persistent and aid in memory retention. The resistance of a spine to erasure is proportional to its volume. 

In physics, a similar relationship exists between mass and frictional resistance: as the mass of an object in motion across a horizontal surface increases, the frictional resistance experienced by the object increases as well. More generally, the magnitude of the friction force is proportional to the object's mass, as demonstrated in the following equation for friction force:
\[F_{friction}=\mu m g,\]
where $\mu$ is the coefficient of friction, $m$ is the mass of the object, and $g$ is the acceleration due to gravity.

Drawing a parallel from these principles in neurology and physics to neural networks, we propose that weights that are relatively large in magnitude should experience more resistance to change, and those that are relatively small in magnitude should experience less resistance to change. In this way, weights of larger magnitude (strong memories) are preserved, while weights of smaller magnitude (weak memories) can be overwritten.

To better understand this, let us first define the \textit{normal update} as the magnitude of the update to a weight when there is no weight friction. When weight friction is present, larger-magnitude weights should have smaller-than-normal updates (more resistance to change), and smaller-magnitude weights should have close-to-normal updates (less resistance to change). We achieve this by modifying the update rule for gradient descent in neural networks. In traditional gradient descent, each weight $w$ is updated by some proportion of the gradient of the loss with respect to the weight itself. With weight friction, we include a multiplicative factor $g(w)$ that is inversely proportional to the magnitude of $w$:
\[w = w-\alpha g(w) \frac{\partial \mathcal{L}}{\partial w},\]
where $\alpha$ is the learning rate. To model this effect, we would like $g(w)$ to resemble a Gaussian curve, with the horizontal spread of the curve scaled by a constant $\mu$ and corresponding to the threshold for a weight's magnitude to be defined as “large.” Note that we can also choose to vary the value of $\mu$ during training.
\begin{figure}[h]
\centering
\begin{minipage}{0.45\textwidth}
    \centering
    \includegraphics[height=1.2in]{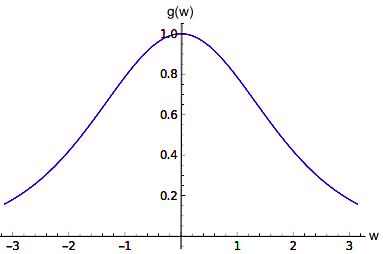}
    \caption{A graph of $g(w)=\frac{4e^x}{(1+e^x)^2}$}
    \label{bellcurve}
\end{minipage}\hfill
\begin{minipage}{0.45\textwidth}
\centering
    \includegraphics[height=1.2in]{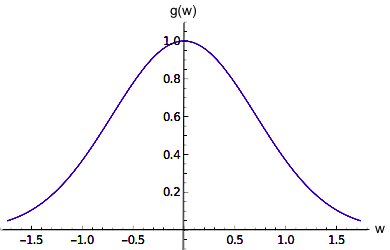}
    \caption{An alternative choice of weight friction function, with $g(w)=e^{-x^2}$}
    \label{altbellcurve}
\end{minipage}
\end{figure}

For the purposes of this paper, we assume \[g(w)=\dfrac{4e^{\mu w}}{(1+e^{\mu w})^2},\] with the graph of this function for $\mu=1$ shown in Figure \ref{bellcurve}. However, note that $g(w)$ can be assigned to any function that behaves similarly. For example, we can also assign $g(w) = e^{-\mu x^2}$, whose graph for $\mu=1$ is shown in Figure \ref{altbellcurve}. To better understand the effect of weight friction in relation to the magnitudes of $w$ and $g(w)$, we can look to Table \ref{tablecases}.

\begin{table}[h]
\captionsetup{justification=centering}
\caption{The behavior of weight friction.}
\centering
\vspace{0.1in}
    \begin{tabular}{ccc}
    \toprule
    $|w|$ & $|g(w)|$ & Weight Friction\\
    \midrule
    Large & Close to 0 & Large\\
    Small & Close to 1 & Small\\\bottomrule
    \end{tabular}
    \label{tablecases}
\end{table}

Notice that the use of weight friction does not modify the loss function itself. Additionally, since $g(w)$ is multiplicative, we can alternatively think of weight friction as a weight-based adaptive learning rate method, whereby the learning rate $\alpha$ is scaled inversely to the magnitude of the weight itself.

%%%%%%%%%%%%%%% CONVERGENCE %%%%%%%%%%%%%%%
\subsection{Convergence}\label{sec3.2}
We show that our weight friction method converges to an optimal solution for the loss function $\mathcal{L}(w)$ at a rate comparable to that of stochastic gradient descent. Let $w^*\in W$ denote the optimal solution to $\min \mathcal{L}(w)$, with\[\argmin_w \mathcal{L}(w) = w^*,\]and let $\nabla_w \mathcal{L}(w_t)$ denote the gradient of the loss function with respect to $w_t$, the parameter vector at the current step $t$. We define the update rule for gradient descent with weight friction to be
\begin{equation}\label{updaterule}
w_{t+1}=w_t-\alpha g(w_t) \nabla_w\mathcal{L}(w_t),
\end{equation} 

with \[g(w_t) =\dfrac{4e^{\mu w_t}}{(1+e^{\mu w_t})^2}\] and $\mu$ a nonnegative value held constant for the purpose of this proof.
To analyze the rate of convergence of gradient descent with weight friction, we would like to find a bound on the regret $R_{\mathcal{L}}(T)$, which is the sum of the differences between $\mathcal{L}(w_t)$ and $\mathcal{L}(w^*)$, at each time step $t\in [1,T]$: \[R_{\mathcal{L}}(T) = \sum_{t=1}^T\left(\mathcal{L}(w_t)- \mathcal{L}(w^*)\right).\]

As weight friction does not alter the loss function $\mathcal{L}$, the method is guaranteed to converge as long as $\mathcal{L}$ satisfies the following two properties \cite{kim2017convergence}:
\begin{itemize}
\item $\mathcal{L}$ is a convex function. That is, \begin{equation}\label{convexity}\mathcal{L}(y)\geq \mathcal{L}(x)+\langle\nabla\mathcal{L}(x), y-x\rangle\quad \forall x, y.\end{equation}
\item $\nabla\mathcal{L}$ is Lipschitz continuous. That is, there exists some constant $L$ such that for all $x$ and $y$, \begin{equation}||\nabla\mathcal{L}(x)-\nabla\mathcal{L}(y) || \leq L||x-y||\label{lipschitz}\end{equation}

\begin{equation}\implies \mathcal{L}(y)\leq \mathcal{L}(x)+\langle\nabla\mathcal{L}(x), y-x\rangle + \frac{L}{2}||y-x||^2 \label{implication}\end{equation}
\end{itemize}

Without loss of generality, for our analysis of convergence, we assume these two properties hold for $\mathcal{L}$. We now present a lemma to be used in a subsequent proof.

\begin{lemma}\label{lemma}
The expression $||w_t||^2-2\langle w_t, w^*\rangle+||w^*||^2 - 2\alpha g \langle \nabla_w\mathcal{L}(w_t), w_t-w^*\rangle + \alpha^2 g^2||\nabla_w\mathcal{L}(w_t)||^2$ is equivalent to $||w_{t+1}-w^*||^2$.
\end{lemma}
\begin{proof}
We begin by separating the fourth term of the initial expression into two terms and apply the commutativity of the dot product to obtain
\begin{multline*}
    ||w_t||^2-2\langle w_t, w^*\rangle+||w^*||^2 - 2\alpha g \langle  w_t, \nabla_w\mathcal{L}(w_t)\rangle + 2\alpha g \langle  \nabla_w\mathcal{L}(w_t), w^*\rangle+ \alpha^2 g^2||\nabla_w\mathcal{L}(w_t)||^2.
\end{multline*}

Next, we rearrange by applying the associativity of the dot product over multiplication by the scalar $\alpha g$, and the expression becomes:
\begin{multline*}
    ||w_t||^2-2 \langle  w_t, \alpha g \nabla_w\mathcal{L}(w_t)\rangle + ||\alpha g \nabla_w\mathcal{L}(w_t)||^2-2\langle w_t, w^*\rangle- 2\langle  - \alpha g \nabla_w\mathcal{L}(w_t), w^*\rangle+||w^*||^2.
\end{multline*}

Now observe that for any two vectors $a$ and $b$, we have\vspace{-1mm} \[||a-b||^2=||a||^2-2\langle a,b\rangle+||b||^2.\]

We can apply this to simplify the previous expression, which becomes\vspace{-1mm} \[||w_t-\alpha g\nabla_w\mathcal{L}(w_t)||^2-2\langle w_t-\alpha g\nabla_w\mathcal{L}(w_t), w^*\rangle +||w^*||^2.\]

This is equivalent to \[||(w_t-\alpha g\nabla_w\mathcal{L}(w_t))-w^*||^2.\] Finally, we complete the proof by substituting $w_t-\alpha g\nabla_w\mathcal{L}(w_t)=w_{t+1}$, to obtain as desired: 
\[||w_{t+1}-w^*||^2.\]
\end{proof}
We now present a key theorem on the rate of convergence of training with weight friction.
\begin{theorem}
If $\mathcal{L}(w)$ is convex and its gradient $\nabla_w\mathcal{L}(w)$ is $L$-Lipschitz continuous, then for $\alpha\in \left(0,\frac{1}{L}\right]$ and $\mu\in \mathbb{R}_{++}$, the sequence ${w_t}$ generated by Equation \ref{updaterule} satisfies \[R_{\mathcal{L}}(T) = \mathcal{O}\left(|| w_1-w^*||^2\right).\]
\end{theorem}
\begin{proof}
We begin by substituting $w_{t+1}$ and $w_t$ for $y$ and $x$, respectively, into Equation \ref{implication}:
\begin{equation*}
\mathcal{L}(w_{t+1})\leq\mathcal{L}(w_t)+\langle\nabla_w\mathcal{L}(w_t), w_{t+1}-w_t\rangle +\frac{L}{2}||w_{t+1}-w_t||^2
\end{equation*}

We can expand this expression by substituting $w_{t+1}-w_t = -\alpha g(w_t) \nabla_w\mathcal{L}(w_t):$
\begin{equation*}
\begin{split}
    \mathcal{L}(w_{t+1})&\leq \mathcal{L}(w_t)+\langle \nabla_w\mathcal{L}(w_t),-\alpha g(w_t) \nabla_w\mathcal{L}(w_t) \rangle +\frac{L}{2}||-\alpha g(w_t) \nabla_w\mathcal{L}(w_t)||^2\\
    &= \mathcal{L}(w_t)-\alpha g(w_t) ||\nabla_w\mathcal{L}(w_t)||^2 +\frac{L}{2}\left(\alpha g(w_t)\right)^2||\nabla_w\mathcal{L}(w_t)||^2\\
    &= \mathcal{L}(w_t)-\alpha g(w_t)\left(1-\frac{L}{2}\alpha g(w_t)\right)||\nabla_w\mathcal{L}(w_t)||^2
\end{split}
\end{equation*}

Since we assume $\alpha\in\left(0,\frac{1}{L}\right]$, we know $\alpha\leq\frac{1}{L}$. Thus, we can substitute $\frac{1}{L}$ for $\alpha$ while still preserving the inequality:
\begin{equation}\label{smiley}
\begin{split}
     \mathcal{L}(w_{t+1})&\leq \mathcal{L}(w_t) - \alpha g(w_t)\left(1-\frac{1}{2} g(w_t)\right)||\nabla_w\mathcal{L}(w_t)||^2.
\end{split}
\end{equation}

We also know that \[g(w) = \dfrac{4e^{\mu w_t}}{(1+e^{\mu w_t})^2}\leq 1 \quad \forall w \text{ and } \forall \mu>0\] and thus similarly substitute $g(w)=1$ while preserving the inequality. Equation \ref{smiley} becomes
\begin{equation}\label{booface}
    \mathcal{L}(w_{t+1})\leq \mathcal{L}(w_t) - \alpha g(w_t)\frac{1}{2}||\nabla_w\mathcal{L}(w_t)||^2.
\end{equation}

In subsequent steps of this proof we denote $g(w_t)$ as $g$.
Now we know because $\mathcal{L}$ is convex that \[\mathcal{L}(w^*)\geq \mathcal{L}(w_t)+\langle\nabla_w\mathcal{L}(w_t),w^*-w_t\rangle.\]

Rearranging this expression gives
\[\mathcal{L}(w_t)\leq \mathcal{L}(w^*)+\langle\nabla_w\mathcal{L}(w_t),w_t-w^*\rangle,\]

and substituting the right-hand side of this inequality for $\mathcal{L}(w_t)$ in Equation \ref{booface} gives
\begin{equation*}
\begin{split}
\mathcal{L}(w_{t+1}) &\leq \mathcal{L}(w^*)+\langle\nabla_w\mathcal{L}(w_t),w_t-w^*\rangle - \frac{\alpha g}{2}||\nabla_w\mathcal{L}(w_t)||^2\\
&=\mathcal{L}(w^*)+\langle\nabla_w\mathcal{L}(w_t),w_t-w^*\rangle -\frac{\alpha g}{2}||\nabla_w\mathcal{L}(w_t)||^2 + \frac{1}{2\alpha g}(||w_t-w^*||^2-||w_t-w^*||^2)\\
&=\mathcal{L}(w^*)+\langle\nabla_w\mathcal{L}(w_t),w_t-w^*\rangle-\frac{\alpha g}{2}||\nabla_w\mathcal{L}(w_t)||^2 + \frac{1}{2\alpha g}(||w_t-w^*||^2\\&\qquad\qquad\quad\qquad\qquad\qquad\qquad\qquad\qquad\qquad\qquad\qquad-
(||w_t||^2-2\langle w_t, w^*\rangle+ ||w^*||^2)).
\end{split}
\end{equation*}

Expanding the three terms that follow $\mathcal{L}(w^*)$ in the last expression and rearranging, we have  \begin{equation*}
\begin{split}
\mathcal{L}&(w_{t+1}) \leq \mathcal{L}(w^*)+\frac{1}{2\alpha g}(||w_t-w^*||^2-(||w_t||^2-2\langle w_t,w^*\rangle +||w^*||^2 -2\alpha g\langle \nabla_w\mathcal{L}(w_t),w_t-w^*\rangle \\&\qquad\qquad\qquad+\alpha^2g^2||\nabla_w\mathcal{L}(w_t)||^2)).
\end{split}
\end{equation*}

We apply Lemma \ref{lemma} to simplify the left-hand side of this expression, which yields \begin{equation}\label{regretterm}
\begin{split}
&\mathcal{L}(w_{t+1}) \leq \mathcal{L}(w^*)+\frac{1}{2\alpha g}\left(||w_t-w^*||^2-||w_{t+1}-w^*||^2\right)\nonumber\\
\implies &\mathcal{L}(w_{t+1}) -\mathcal{L}(w^*) \leq \frac{1}{2\alpha g}\left(||w_t-w^*||^2-||w_{t+1}-w^*||^2\right).
\end{split}
\end{equation}

Lastly, summing the left-hand side of Equation \ref{regretterm} for all $t$, we are able to derive a bound on the regret $R_{\mathcal{L}}(T)$, as desired:
\begin{align*}
    R_{\mathcal{L}}(T)&= \sum_{t=1}^T\left[\mathcal{L}(w_{t}) -\mathcal{L}(w^*) \right]\\ &\leq \frac{1}{2\alpha g}\sum_{t=1}^T\left[||w_t-w^*||^2-||w_{t+1}-w^*||^2\right]\\
    &= \frac{1}{2\alpha g}\left[||w_1-w^*||^2-||w_{T+1}-w^*||^2\right]\\
    &\leq \frac{1}{2\alpha g}||w_1-w^*||^2
\end{align*}

Thus, we have \[R_{\mathcal{L}}(T)\leq\frac{1}{2\alpha g}||w_1-w^*||^2,\] which means that training with weight friction converges at a rate on the order of $\mathcal{O}\left(||w_1-w^*||^2\right)$, similar to that of training with stochastic gradient descent \cite{kim2017convergence}.
\end{proof}

%%%%%%%%%%%%%%% EMPIRICAL EVALUATION %%%%%%%%%%%%%%%
\subsection{Empirical analysis}\label{sec3.3}

We defined three experimental settings to analyze the effects of weight friction.

Settings 1 and 2 were based on the image classification benchmark datasets MNIST and Fashion-MNIST ~\cite{mnistcitation,xiao2017fashion}. MNIST consists of 70,000 grayscale images sized 28x28, each showing a single, centered handwritten digit from one of 10 classes (0-9). Fashion-MNIST consists of 70,000 grayscale images sized 28x28, each showing a single, centered article of clothing from one of 10 classes. For each dataset, we used the provided 10,000-example test set and randomly sampled the remaining examples according to an 80/20 train/validation split.

In Setting 1, a neural network was first trained on the MNIST dataset to perform the task of handwritten digit classification. Then, the same neural network continued to train on the Fashion-MNIST dataset to perform the task of object classification. Lastly, with no further training, we re-evaluated the model's performance on the first task on MNIST. 

Setting 2 was identical to Setting 1, but with models trained initially on Fashion-MNIST, then trained on MNIST second, and finally re-evaluated on Fashion-MNIST. 

Setting 3 was based on Permuted MNIST \cite{goodfellow2013empirical}, a variation of MNIST. With Permuted MNIST, new tasks of comparable difficulty to the original MNIST classification task are created by permuting the pixels of every image based on a randomly generated permutation. Similar to Settings 1 and 2, in Setting 3, a neural network was trained successively on ten Permuted MNIST tasks, and the model’s performance on previous and current tasks was re-evaluated after training on each new task. Training and test sets were derived from the original 60,000- and 10,000-example training and test sets. No validation data was used in this setting; cross-validation was used for parameter tuning.

In Settings 1-2, model architectures were feedforward neural networks consisting of a 784-neuron input layer followed by three 256-neuron hidden layers with ReLU activation and one 10-neuron output layer with softmax activation. No dropout or regularization was used, and the learning rate was 0.01. Each model was trained until convergence at 50 epochs on the first task and 100 epochs on the second task. For both settings, we trained models both with and without weight friction. Models trained without weight friction used Adam optimization and served as a baseline for comparison. 

In Setting 3, model architectures were feedforward neural networks consisting of a 784-neuron input layer, two 256-neuron hidden layers with ReLU activation, and one 10-neuron output layer with softmax activation, with no dropout and no regularization and trained for 5,000 epochs per task. We compared models trained with weight friction (WF) to several models trained with other continual learning methods (EWC, PNN, A-GEM) and to a baseline model trained without weight friction and with Adam optimization (VAN). For VAN, EWC, PNN, and A-GEM, all hyperparameters were identical to those used by \cite{chaudhry2018efficient} except learning rate, which was tuned to 0.001.

In all settings, models trained with weight friction were optimized over $\mu$ by gridsearch. For all models, weight friction was applied only to tasks following the first; no weight friction was applied to training on the first task in any setting. We used test set accuracy as a metric to evaluate model performance and averaged the results of 10 random initializations to account for variability. All model parameters were initialized with Xavier initialization~\cite{glorot2010understanding}.

% \begin{figure}[h]
% \hfill
% \subfigure[]{\includegraphics[width=0.47\linewidth]{set1.png}}
% \hfill
% \subfigure[]{\includegraphics[width=0.47\linewidth]{set2.png}}
% \hfill
% \caption{Average test set accuracies for (a) Setting 1 and (b) Setting 2.}
% \end{figure}

\begin{figure}[h]
\centering
\begin{minipage}[t]{0.47\textwidth}
\centering
\includegraphics[width=0.9\linewidth]{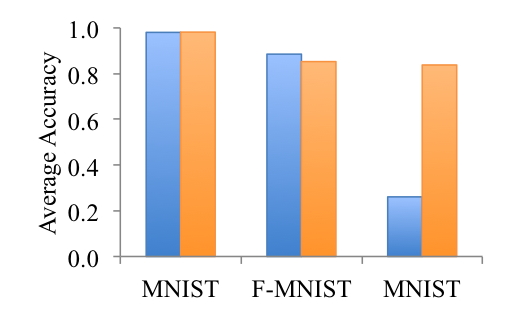}
\caption{Average accuracies for setting 1.}
\label{set1}
\end{minipage}\hfill
\begin{minipage}[t]{0.47\textwidth}
\centering
\includegraphics[width=0.9\linewidth]{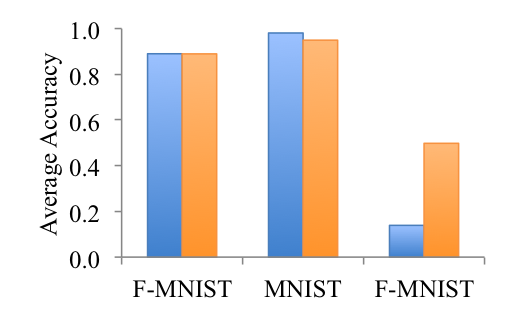}
\caption{Average accuracies for setting 2.}
\label{set2}
\end{minipage}
\end{figure}

\begin{figure}[h]
\centering
\begin{minipage}[t]{0.47\textwidth}
\centering
\includegraphics[width=\linewidth]{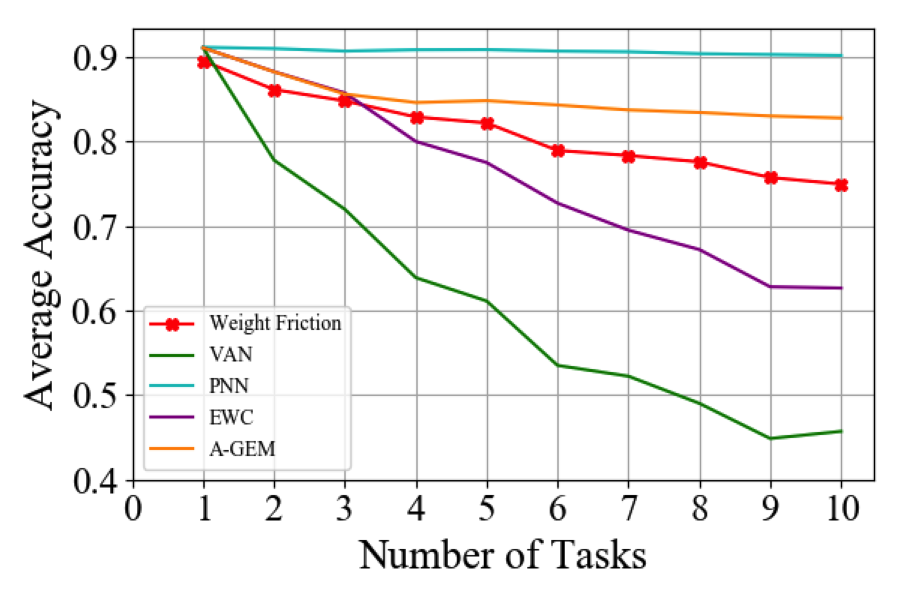}
\caption{Average accuracies for setting 3, computed as $\frac{1}{n}\sum_{k=1}^{n} a_k$, where $n$ is the number of tasks and $a_k$ is the test set accuracy on task $k$.}
\label{accuracies}
\end{minipage}\hfill
\begin{minipage}[t]{0.47\textwidth}
\centering
\includegraphics[width=\linewidth]{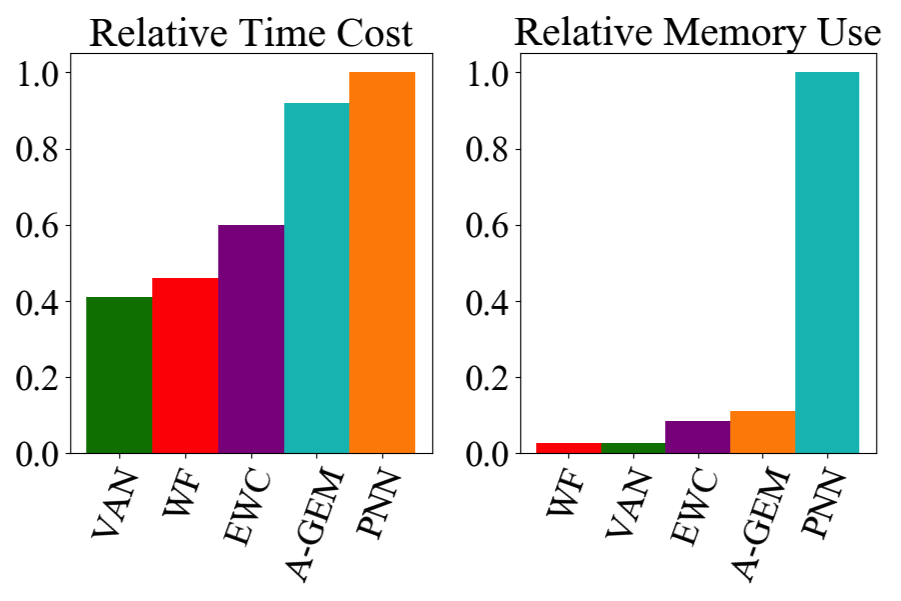}
\caption{Comparison of time and memory costs, relative to the method with the greatest respective cost (which is given a reference value of 1).}
\label{timemem}
\end{minipage}
\end{figure}

As shown in Figures \ref{set1} and \ref{set2}, the baseline models trained without weight friction (blue bars) exhibited catastrophic forgetting. Average test accuracy for the first task decreased from 98.08\% to 26.09\% in Setting 1 and from 89.00\% to 13.89\% in Setting 2 following training on the second task. In comparison, the models trained with weight friction (orange bars) were better able to learn and remember representations across task domains. Specifically, the average final test accuracy on the first task rose from 26.09\% to 83.82\% in Setting 1 and from 13.89\% to 49.81\% in Setting 2. At the same time, the average final test accuracy achieved with weight friction for the second task remained nearly the same as the baseline in both settings. This evidences that weight friction enables neural networks to overcome catastrophic forgetting and learn representations that facilitate continual learning.

Differences in the results of Settings 1 and 2 further suggest that the order in which tasks are learned impacts the efficacy of weight friction. Specifically, in both settings, the model ultimately learned to classify images from both MNIST and Fashion-MNIST. Yet training on MNIST before Fashion-MNIST (Setting 1) led to better results overall, with the model achieving average accuracies of 85.29\% and 83.82\% on Fashion-MNIST and MNIST, respectively, in Setting 1 versus the 49.81\% and 94.95\% achieved in Setting 2. It is possible that weight friction allows neural networks to learn representations on simpler tasks that are transferred when learning more complex tasks. This potentially reflects the progression of human learning from simple to complex tasks.

The performance results of Setting 3 show that weight friction is comparable to existing approaches. From Figure \ref{accuracies}, it is evident that weight friction is not limited to operate over a small number of task domains and can be applied to a large number of tasks without compromising model accuracy. In terms of average model accuracy, weight friction outperforms EWC as more tasks are learned. Figure \ref{timemem} indicates that weight friction results in a much lower computation time and memory cost relative to EWC, PNNs, and A-GEM. It is important to note that training a neural network with weight friction resulted in a negligible increase in memory use from vanilla training without weight friction. Furthermore, in terms of computation time during training, weight friction was 2.16 times faster than A-GEM, 1.98 times faster than PNNs, and 1.29 times faster than EWC. Weight friction yielded even greater improvements in terms of memory cost during training, with memory use 35.71 times lower than PNNs, 3.57 times lower than A-GEM, and 3.04 times lower than EWC. While PNNs and A-GEM achieved highest accuracy, they also resulted in the worst efficiency, with the highest memory and computation time costs, respectively. In comparison, weight friction achieved greatest efficiency and relatively high accuracy, which indicates that weight friction offers one of the best tradeoffs between accuracy and efficiency.

%%%%%%%%%%%%%%%%%%%%%%%%%%%%%
%% Discussion
%%%%%%%%%%%%%%%%%%%%%%%%%%%%%
\section{Discussion}\label{sec4}

The implications of our results are significant in comparison to those of existing approaches. In particular, weight friction does not suffer from common limitations of current methods for facilitating continual learning. For instance, unlike Progressive Neural Networks, weight friction does not significantly increase computational complexity or the number of model parameters that must be learned, as it does not require the instantiation of a new neural network for each task of interest~\cite{rusu2016progressive}. Unlike GEM and A-GEM, it does not require the use of an episodic memory during training \cite{lopez2017gradient,chaudhry2018efficient}. Unlike importance factor-based methods, it does not require prior knowledge or supervision regarding which weights should be preserved, which simplifies training. Furthermore, unlike methods that seek to maximize domain confusion, weight friction is not restricted by design to operate over a limited set of task domains~\cite{tzeng2014deep}.

The benefits of weight friction are potentially magnified when applied to other types of neural networks. In particular, as the weight friction mechanism depends only on a modification to the update rule for gradient descent, it is inherently applicable to any neural network architecture. This therefore allows us to harness weight friction to overcome catastrophic forgetting and facilitate continual learning in other types of neural networks.

%%%%%%%%%%%%%%%%%%%%%%%%%%%%%
%% Conclusions
%%%%%%%%%%%%%%%%%%%%%%%%%%%%%
\section{Conclusions}\label{sec5}
In this research, we addressed the problem of catastrophic forgetting in neural networks, which hinders continual learning. Specifically, we drew from principles in neurology and physics to develop the concept of weight friction as a new learning method. We showed that training with weight friction converges at a rate comparable to that of stochastic gradient descent. We showed empirically that neural networks trained with weight friction can potentially learn representations shared across data domains. We further discussed the various benefits of weight friction as an approach that is less complex, more efficient, and more widely applicable in comparison to existing methods. Ultimately, this research takes a step toward building neural networks with continuous learning ability.

There are several areas we would like to pursue in our future work. First, we seek to derive a bound on the information capacity of a neural network trained with weight friction. Namely, how many tasks and of what complexity can such a neural network learn while maintaining a reasonable level of performance for each task? Another avenue for investigation is the application of weight friction to other neural network architectures, including convolutional and recurrent neural networks. We also hope to study the effect of weight friction function $g(w)$ selection on performance. Lastly, we would like to analyze the effect of learning algorithms that combine weight friction, momentum, regularization, and/or normalization.

%%%%%%%%%%%%%%%%%%%%%%%%%%%%%
%% REFERENCES
%%%%%%%%%%%%%%%%%%%%%%%%%%%%%
\printbibliography
% \section*{References}
% \small
% \bibliography{NeuRIPS2019/paper}

\end{document}